\documentclass[numbers,preprint,review,12pt]{elsarticle}

\usepackage{hyperref}

\journal{Neurocomputing}

\usepackage{graphicx}
\usepackage{multirow}
\usepackage{times}
\usepackage{amsmath,amsthm}
\usepackage{amssymb}
\usepackage{amsfonts}
\usepackage{algorithm}
\usepackage{mathrsfs}
\usepackage{booktabs} 
\usepackage{color}
\usepackage{hyperref} 
\usepackage{url}
\usepackage{caption} 
\usepackage{algpseudocode}
\usepackage{pifont}
\usepackage{bm}
\usepackage{comment}
\usepackage{wasysym}
\usepackage{soul}
\usepackage{CJKutf8}

\begin{document}

\begin{frontmatter}



\title{Endowing Embodied Agents with Spatial Reasoning Capabilities for Vision-and-Language Navigation}

\author[swufe-a]{Qianqian~Bai}
\author[swufe-a]{Zhongpu Chen}
\author[swufe-a]{Ling Luo}
\author[swufe-a]{Huaming Du}
\author[swufe-b]{Yuqian Lei}
\author[swufe-c]{Ziyun Jiao}
\cortext[mycorrespondingauthor]{Corresponding author}\ead{zhaoyu@swufe.edu.cn(Yu Zhao);7220323@ustc.edu.cn}
\author[swufe-a]{Yu Zhao}


\address[swufe-a]{Southwestern University of Finance and Economics, Chengdu, 611130, China}
\address[swufe-b]{Department of Informatics,Universitat Hamburg, Hamburg, Germany \\}
\address[swufe-c]{University of Electronic Science and Technology, Chengdu, China \\}

\begin{abstract}
Enhancing the spatial perception capabilities of mobile robots is crucial for achieving embodied VLN. Although significant progress has been made in simulated environments, directly transferring these capabilities to real-world scenarios often results in severe hallucination phenomena, causing robots to lose effective spatial awareness. To address this issue, we propose BrainNav, a bio-inspired spatial cognitive navigation framework inspired by biological spatial cognition theories and cognitive map theory. BrainNav integrates dual-map (coordinate map and topological map) and dual-orientation (relative orientation and absolute orientation) strategies, enabling real-time navigation through dynamic scene capture and path planning. Its five core modules—Hippocampal Memory Hub, Visual Cortex Perception Engine, Parietal Spatial Constructor, Prefrontal Decision Center, and Cerebellar Motion Execution Unit—mimic biological cognitive functions to reduce spatial hallucinations and enhance adaptability. Validated in a zero-shot real-world lab environment using the Limo Pro robot, BrainNav outperforms existing state-of-the-art VLN methods in a variety of aspects, without requiring any fine-tuning.
\end{abstract}




\end{frontmatter}



\section{Introduction}

Embodied VLN \cite{anderson2018vision} is a crucial research direction in artificial intelligence, aiming to enable intelligent agents to navigate complex environments through natural language instructions. Although significant progress has been made in VLN research based on simulated environments in recent years, when transferring models from simulated to real-world environments, agents often face severe spatial hallucination \cite{williams2024EasyProblems,gao2024vision} issues. This refers to the inconsistency between the agent's perception of the environment during navigation and the actual environment, leading to a loss of effective spatial understanding and ultimately failure to complete navigation tasks. 

The primary limitation of current VLN tasks lies in agents' over-reliance on single cue systems and mapping frameworks, lacking multi-level environmental modeling and dynamic updating capabilities \cite{macdonald2024language,ahn2024autort,huang2023instruct2act}. Firstly, single cue systems only retain target information while neglecting environmental details, leading to random direction choices and path repetition in similar scenarios. Secondly, existing methods depend solely on either global maps or local path planning strategies \cite{chen-etal-2024-mapgpt,rana2023sayplan}, failing to effectively encode real-world spatial relationships. This results in delayed updates of spatial representations, compromising navigation accuracy and adaptability.

\begin{figure}[!t]
  \includegraphics[width=1\textwidth]{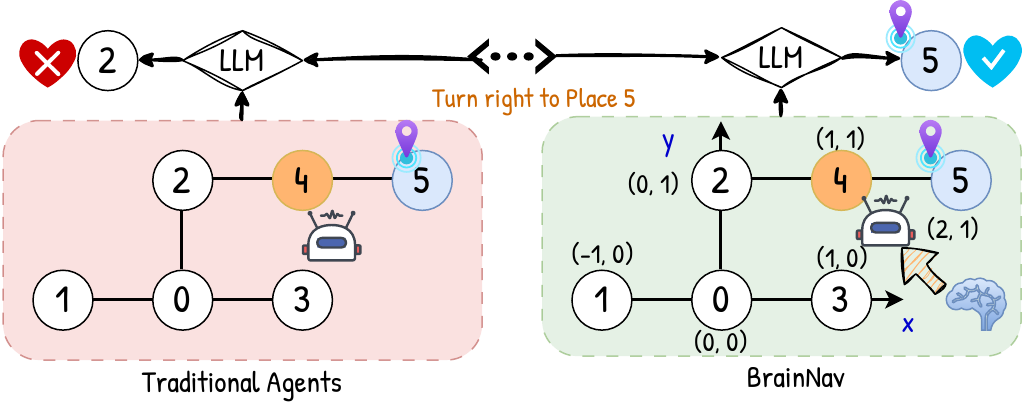}
  \caption{Comparison of the reasoning processes between traditional GPT-based navigation agents and BrainNav in real-world environments. Traditional agents rely solely on local action spaces and are prone to spatial illusions due to a lack of global understanding of spatial structures. In contrast, BrainNav draws inspiration from biological mechanisms, introducing a dual-map system (coordinate and topological maps) and dual-orientation strategies (relative and absolute) to help the agent build spatial relational awareness, enabling precise reorientation and global path planning.}
  \label{fig:mind}
\end{figure}

To address these challenges, we propose BrainNav, a biomimetic navigation framework inspired by biological spatial cognition theories and cognitive mapping mechanisms \cite{o1991biologically}. As illustrated in Fig.~\ref{fig:mind}, compared to traditional navigation systems, biological navigation systems demonstrate superior environmental adaptability by efficiently constructing and updating spatial cognition in dynamic scenarios. Neuroscience research reveals that the hippocampus, visual cortex, and parietal lobe collaboratively support biological spatial cognition: the hippocampus encodes spatial relationships through topological maps, the visual system extracts local features, and the parietal lobe integrates and updates environmental information \cite{stachenfeld2017hippocampus,sack2009parietal}.

Specifically, BrainNav simulates the \emph{multi-level spatial representation} mechanism of mammals by employing a dual-map architecture (coordinate map and topological map) to replicate the path integration function of the hippocampus. The coordinate map corresponds to grid cell encoding, providing absolute positional awareness. The topological map serves as a computational implementation of the hippocampal memory hub, leveraging the biological mechanism by which place cells are activated when the agent enters specific locations to construct and continuously update spatial memory, thereby capturing the structural relationships of the environment \cite{kim2019can,moser2008place}. Meanwhile, through \emph{path backtracking} and \emph{memory recall} mechanisms, it effectively mitigates erroneous decisions caused by spatial illusions, enhancing the agent’s navigation stability and adaptability in complex, dynamic environments. Additionally, we introduce a \emph{dual-orientation} strategy (relative and absolute orientation), inspired by the biological characteristics of head-direction cells that sense an individual's head orientation and provide absolute directional information. This strategy integrates the functions of grid cells and place cells, enabling the system to dynamically fuse self-motion cues with environmental perception data \cite{moser2008place,taube1998head}. This design endows BrainNav with dynamic spatial representation updating capabilities akin to biological neural systems, significantly enhancing robustness and adaptability in real-world environments.

We validated BrainNav in a zero-shot real-world laboratory environment using the Limo Pro robot. Experimental results show that BrainNav is compatible with large models without fine-tuning and significantly outperforms existing state-of-the-art (SOTA) Vision-and-Language Navigation methods. The implementation of BrainNav is publicly available at: \url{https://github.com/swufe-agi/BrainNav}.

Our key contributions are threefold:
\begin{itemize}
    \item We propose BrainNav, the first bio-inspired hierarchical decision system, consisting of a hippocampal memory module, a visual cortex perception engine, a parietal space constructor, a prefrontal decision center, and a cerebellar motion execution unit, which improves the agent's spatial perception and decision-making.
    
    \item Specifically, we propose a dual-map dual-orientation navigation framework, integrating bio-inspired spatial cognition theory. It combines coordinate and topological maps for multi-level spatial representation and incorporates both relative and absolute orientation for enhanced spatial awareness.
   
    \item Without fine-tuning, BrainNav has achieved SOTA results across various navigation tasks, significantly outperforming existing methods, fully demonstrating its superiority in real-world environments.
\end{itemize}

\section{Related Work}
\subsection{Vision-and-language Navigation}



VLN \cite{anderson2018vision,qi2020reverie,ku2020room} aims to enable agents to navigate complex environments based on human instructions and visual observations. With the rapid development of LLMs \cite{chiang2023vicuna,zhu2023chatgpt} and Vision-Language Models (VLMs) \cite{liu2023visual,chen2024internvl,zhang2024llama}, VLN research in simulated environments has made significant progress. For example, NaviLLM \cite{zheng2024towards} unifies multiple navigation tasks through pattern-based instructions, DiscussNav \cite{long2024discuss} introduces a collaborative mechanism among domain experts, NavGPT \cite{zhou2024navgpt} integrates LLMs with policy networks in a two-stage text-based framework, while LM-Nav \cite{shah2023lm} and ImagineNav \cite{zhao2024imaginenav} enhance spatial perception and planning capabilities via landmark extraction and future observation generation, respectively. However, these methods primarily rely on simulator-based training and often overlook real-world deployment challenges such as perception noise, dynamic environments, and incomplete maps. Consequently, agent performance drops significantly when transferred to real-world scenarios. Studies \cite{sahoo2024comprehensive, williams2024easy,zhang2023siren,gao2024vision} have revealed limitations in LLMs' spatial reasoning abilities, with agents prone to spatial hallucinations in the real world.

To address these challenges, we propose the BrainNav framework, which enhances agents' environmental understanding through efficient map construction and real-time updates, thereby significantly improving their adaptability and robustness in complex real-world settings.

\subsection{Visual Prompt Engineering}


Visual Prompt Engineering enhances an agent’s perception and understanding by designing effective image-based prompts, thereby improving performance in complex tasks. The introduction of multimodal attention mechanisms~\cite{landi2021multimodal} and cross-modal embedding alignment~\cite{radford2021learning} has inspired new directions in prompt design. For instance, CLIP~\cite{shtedritski2023does} uses visual annotations to support zero-shot image understanding; DPT~\cite{zhou2022learning,yang2023dynamic} proposes dynamic prompt tuning to adapt to varying tasks and environments; Multimodal-CoT~\cite{zhang2023multimodal} integrates textual and visual inputs for enhanced reasoning; and SAA+~\cite{cao2023segment} combines regularized prompts with domain knowledge to improve anomaly localization.

However, existing methods still face limitations in flexibility and adaptability, restricting the full potential of visual information. In embodied navigation tasks, we introduce dynamic elements—such as natural landmarks, path markings, and relative distances—into visual prompt design, enabling agents to better understand unknown environments for more effective localization and navigation.

\subsection{Navigation with Online Mapping}


Navigation with Online Mapping enables embodied agents to perform tasks such as point-goal navigation, object-goal navigation, and image-goal navigation by building and updating maps in real time within unknown environments \cite{wu2022image}. Existing methods are mainly categorized into metric maps \cite{mur2015orb, chaplot2020learning, karkus2021differentiable} and topological maps, based on how the environment is represented. Metric maps use explicit geometry (e.g., coordinate maps) or neural representations to model the environment’s structure and appearance \cite{teed2021droid, dai2022planeslam, sandstrom2023point, sucar2021imap, zhu2022nice, wang2023co}, but suffer from heavy reliance on precise geometry and high storage costs, limiting scalability. Topological maps, which highlight connectivity via node-edge structures, include graph-based \cite{lux2024topograph} and semantic-based topologies \cite{wang2022lightweight}, but are often tailored to specific tasks and require targets in predefined formats (e.g., pose or category labels). In vision-language navigation within real-world environments, how to construct maps from language prompts for planning remains underexplored. 

To address this, we propose a dual-map architecture (coordinate + topological maps) inspired by the hippocampal path integration mechanism. Combined with a dual-orientation strategy (relative + absolute), the system dynamically captures scene images via camera, aligns them with coordinates, and updates multi-level path information in real time for efficient and robust navigation.

\section{Methodology}

\begin{figure}[!h]
    \centering
    \includegraphics[width=1\textwidth]{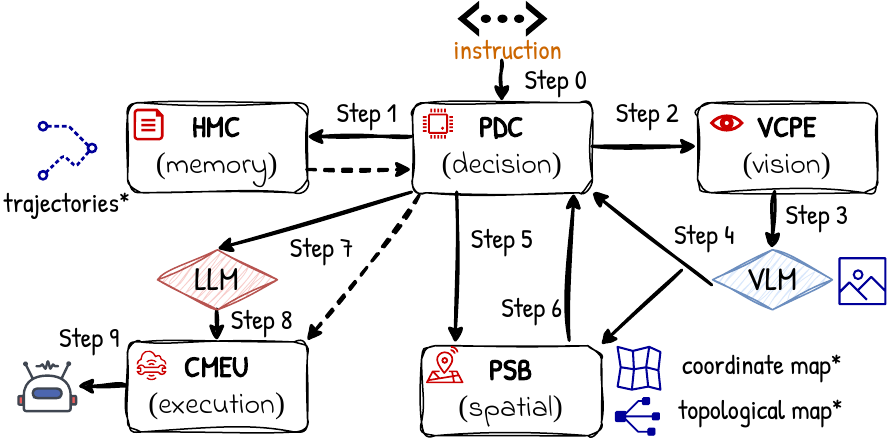} 
    \caption{The overall architecture of BrainNav implements a closed-loop navigation process that spans natural language understanding, perception, memory, planning, and execution.} 
    \label{fig:BrainNav} 
\end{figure}

In this section, we introduce the architectural design of BrainNav. As shown in Figure~\ref{fig:BrainNav}, the system comprises the following modules: the hippocampal memory center for storing and recalling historical information; the visual cortex perception engine for extracting environmental data; the parietal spatial builder for constructing spatial maps; the prefrontal decision center for high-level task planning; and the cerebellum motor execution unit for performing precise actions. 

To be specific, upon receiving an instruction, the Prefrontal Decision Center (\textbf{PDC}) assigns the task (Step 0); the Hippocampal Memory Center (\textbf{HMC}) retrieves historical trajectories (Step 1); the Visual Cortex Perception Engine (\textbf{VCPE}) analyzes environmental images to extract semantic and spatial information (Step 3); and the Parietal Spatial Builder (\textbf{PSB}) updates the map and returns processed information (Step 5). Guided by a large language model, the PDC performs goal reasoning and generates an action plan along with motor commands (Step 4, Step 6, and Step 7), which are then executed by the Cerebellar Motion Execution Unit (\textbf{CMEU}) at Step 8 and 9. During each execution cycle, the system continuously updates the variables annotated with asterisks, including trajectories*, topological map*, and coordinate map*. Note that the dashed line indicates that when reusable historical trajectories are available, the system can bypass high-level reasoning and directly retrieve past trajectories to enable fast decision-making. These modules work collaboratively to enable efficient navigation and decision-making in complex environments. Details of prompt design are provided in the appendix.

\subsection{Hippocampal Memory Center}
\label{sec:hmc}


\subsubsection{Memory Recall in Decision-Making}

The hippocampal memory center plays a key role in the agent’s decision-making process. Upon receiving task instructions (Step 0), the system retrieves the topological map and historical trajectories to assist the decision center in planning (Step 1). The topological map encodes location connectivity in the form of a dictionary. At the same time, historical navigation data—such as previously executed actions and their corresponding images—are accessed. This allows the agent to recall previously traversed paths relevant to the current task, reducing perception errors and improving decision stability.

\subsubsection{Path Backtracking and Experience Reuse}

Another function of the hippocampal memory center is to support path backtracking and reuse of prior experiences. When the agent re-encounters the same task or location, the system can reuse historical trajectories. After each instruction is executed, the system updates navigation history, including motion paths, path choices, and environmental observations. The historical action sequence \( H_t \) is defined as:

\begin{equation}
    H_t = \{a_1, a_2, \dots, a_k\},
\end{equation}
where \( a_k \) denotes an individual action, and \( H_t \) represents the action sequence at time step \( t \). A single time step may contain multiple consecutive actions, reflecting adjustments made by the agent within the same decision cycle based on environmental feedback. The system searches the topological map using graph algorithms to find backtracking paths, prioritizing those with the fewest steps. Once a valid path is found, the corresponding historical trajectory is loaded and executed to guide the agent efficiently back to the target, avoiding redundant exploration. Through the combined mechanisms of memory recall and path backtracking, the hippocampal memory center reduces decision errors caused by spatial illusions and enhances the agent’s adaptability and navigation performance in complex, dynamic real-world environments.

\subsection{Visual Cortex Perception Engine}
\label{sec:vcpe}

The \textbf{VCPE} leverages large language models to analyze images, identify obstacles, traversable areas, and landmarks, and generate semantic descriptions of the environment to provide perceptual input for the decision-making module (Step 3).

\subsubsection{Hierarchical Decision-Making Mechanism}

Conventional approaches (e.g., MapGPT \cite{chen-etal-2024-mapgpt}) typically employ large models as a single expert to process multimodal inputs and generate navigation decisions. While these methods perform well in simulated environments, they are prone to spatial hallucinations in real-world scenarios, often resulting in repeated paths or collisions. This is mainly because simulated environments simplify navigation tasks—if a goal exists or is likely to appear in the next step, the agent can directly shift its view to that location. In contrast, navigation in real-world environments requires precise directional control and action execution for movement between positions. Moreover, images from similar viewpoints in real environments are often difficult to distinguish. Single-expert systems tend to overlook detailed environmental analysis, leading to random direction choices, obstacle collisions, or redundant exploration.

To address these issues, we introduce a hierarchical decision-making mechanism that decomposes the task into multiple submodules. A vision-language model is used to perform structured multi-stage processing of images, including: (1) object recognition and semantic list generation; (2) traversable area labeling; and (3) relative distance estimation. This mechanism enhances scene understanding and improves navigation stability.

\subsubsection{Image Processing and Multimodal Information Extraction}

The \textbf{VCPE} captures images from four directions around the agent, identifies key objects, extracts traversable paths (e.g., “walkable corridor”), and generates path node sequences to represent navigable areas. At the same time, the system estimates the relative distance between the agent and target objects, providing spatial references for path planning. The structured semantic information generated integrates image content with environmental understanding, significantly improving navigation efficiency (Step 4).

\subsection{Parietal Spatial Builder}
\label{sec:psb}

The \textbf{PSB} integrates coordinate and topological maps for environmental modeling. The coordinate map provides precise localization, while the topological map captures spatial connectivity. Together, they support perception and navigation (Step 5).

\subsubsection{Construction and Update of the Coordinate Map}

The coordinate map is built based on the agent’s movement. After each step, the system updates four candidate points around the current location. Given the current position $(x_c, y_c)$, the candidates are computed as:

\begin{equation}
    (x_i, y_i) = (x_c + \Delta x_i, y_c + \Delta y_i), \quad i \in \{0,1,2,3\},
\end{equation}

where $\Delta x_i, \Delta y_i$ depend on the agent’s orientation $\theta$. Orientation updates follow:

\begin{equation}
    \theta_i = (\theta + i \times 90^\circ) \bmod 360^\circ
\end{equation}

Each candidate point is assigned a unique ID that increases dynamically during exploration.

\subsubsection{Construction and Update of the Topological Map}

The topological map is a graph $G = (V, E)$, where nodes represent key locations and edges denote navigable paths. When the agent reaches a new node $v_i$, the adjacency list is updated:

\begin{equation}
    E = E \cup \{(v_c, v_i)\}
\end{equation}

If $v_i$ is not yet in the graph, it is added to $V$. The path mapping is also updated:

\begin{equation}
    \mathrm{roadgraph}[v_i] = \mathrm{roadgraph}[v_i] \cup \{v_j\}
\end{equation}

 During navigation, the system selects optimal candidate points based on topological connectivity to improve path efficiency.

\subsection{Prefrontal Decision Center}
\label{sec:pdc}

The \textbf{PDC} simulates the cognitive functions of the human prefrontal cortex by integrating information from the Hippocampal Memory Hub, Visual Perception, Spatial Mapping, and Motion Execution units (Step 4 and 6). The \textbf{PDC} uses a large language model to fuse these multimodal inputs and generates a comprehensive action plan with three core outputs: task-related decisions, subsequent action sequences, and precise motion command lists. This module enables the agent to make accurate decisions and execute efficiently in complex dynamic environments, bridging cognitive planning and physical actions, similar to the coordination of perception, memory, and movement in the human brain.

\subsection{Cerebellum Motor Execution Unit}
\label{sec:cmeu}

The \textbf{CMEU} simulates the cerebellum’s motor coordination function, translating high-level commands from the Prefrontal Decision Center into precise low-level robotic actions (Step 7).

\subsubsection{Macro Action Conversion}

The \textbf{CMEU} uses predefined macro actions to convert high-level commands such as “move forward,” “turn left,” and “backtrack” into executable control signals(Step 8). For example, the “move forward” command is translated into motor control signals that drive movement by moving forward, while the “turn left” command adjusts the robot’s rotation angle by turning left. The standardized design of macro actions enhances execution efficiency, simplifies parsing, and ensures precise movement across various environments (Step 9).

\subsubsection{Absolute and Relative Orientation}
The \textbf{CMEU}  primarily uses relative orientation (e.g., “left” indicates a 90° counterclockwise turn from the current direction) for general motion, while incorporating absolute orientation (relative to a global coordinate system, where 0° points North) during backtracking to reduce spatial hallucinations. It calculates the coordinate differences:

\begin{equation}
    \Delta x = x_{\mathrm{target}} - x_c, \quad \Delta y = y_{\mathrm{target}} - y_c,
\end{equation}
then updates the heading:

\begin{equation}
    H_{\mathrm{new}} = (H_{\mathrm{current}} + \Delta \theta) \bmod 360^\circ,
\end{equation}
where $\Delta \theta$ is determined by the movement direction. This mechanism allows the robot to progressively follow the backtracking path, adjust its orientation, and ultimately reach the target viewpoint.

\section{Experiment}

\subsection{Experimental Setting}

\subsubsection{Robot Setup}


This study uses the Sunling Limo PRO mobile robot equipped with an Orbbec DaBai camera to capture RGB images of the current view. We implemented five basic actions: “forward” moves the robot 0.5 cm ahead; “backward” rotates the robot base 180° and moves 0.5 cm backward; “turn left” and “turn right” rotate the robot base by 90° respectively; “stop” keeps the robot stationary. Notably, our experimental framework relies solely on RGB images without any algorithms or input from depth cameras, LiDAR, or other sensors, making it easy to deploy in real-world environments. Additionally, BrainNav relies on GPT-4o for high-level reasoning and environmental understanding to guide navigation decisions and interpret sensory input.


\subsubsection{Baselines}






We selected MapGPT~\cite{chen-etal-2024-mapgpt} as our baseline mainly because it aligns closely with our research goals. MapGPT converts map topological relationships into textual prompts, constructing map information that guides GPT in global exploration. Through an adaptive planning mechanism, it activates GPT’s multi-step path planning ability to systematically explore multiple potential targets for VLN tasks~\cite{krantz2020beyond}. In comparison, ViNT~\cite{shah2023vint}, as an end-to-end visual navigation model, lacks environment map modeling and long-term planning, limiting its performance in complex tasks; SayPlan~\cite{rana2023sayplan} relies on costly 3D scene graph construction and classical path planners, increasing system complexity and deployment difficulty; VLMaps~\cite{huang23vlmaps} fuse pretrained visual-language features with 3D reconstruction to achieve precise language indexing but heavily depend on 3D reconstruction hardware, limiting practical deployment; RoboMatrix~\cite{mao2024robomatrix} uses a skill-centric hierarchical framework and unified vision-language-action model to demonstrate strong generalization in open-world tasks, focusing on skill composition and task execution, but does not support efficient conversion of map topology into language prompts or multi-step planning. Overall, MapGPT is the only model that effectively converts map topology into language prompts, supports multi-step planning without extra modules, and best fits our baseline needs.
However, MapGPT’s experiments are limited to simulated environments and based on idealized assumptions, restricting its real-world application. In contrast, our method has two key advantages: (1) no need for simulator training; and (2) like MapGPT, no additional fine-tuning in real environments. These give BrainNav significant practical and generalization advantages.

We evaluate our method using four standard VLN metrics~\cite{krantz2022instance}. Trajectory Length (\textbf{TL}) measures the average steps to complete a task; Navigation Error (\textbf{NE}) is the average distance in meters between the robot’s stopping point and the goal; Success Rate (\textbf{SR}) indicates the proportion of trials where the robot stops within 1 meter of the target, balancing precision and spatial scale; Success weighted by Path Length (\textbf{SPL}) measures navigation efficiency, defined as:

\begin{equation}
    \mathrm{SPL} = \frac{1}{N} \sum_{i=1}^N S_i \cdot \frac{l_i}{\max(p_i, l_i)},
\end{equation}
where $N$ is the total number of trials, $S_i$ is the success indicator for trial $i$ (1 for success, 0 for failure), $l_i$ is the shortest path length from start to goal, and $p_i$ is the actual path length traveled. Higher SPL indicates better navigation efficiency.

\subsubsection{Environment Setup}

The experimental scenarios cover various indoor environments, including laboratories, corridors, meeting rooms, and open areas, to test the adaptability of the proposed method to different spatial layouts and semantics.

    
    
    

\subsection{Experimental Results}

A total of 200 instructions of varying difficulty were designed, split evenly into 100 simple and 100 complex instructions. Each instruction was repeated three times. To maintain real-world conditions, obstacles were randomly added along the robot’s path to the target, evaluating navigation performance in complex environments.

\subsubsection{Simple Instruction Tasks}

Simple instructions include path navigation and target search. The former focuses on basic path planning, while the latter tests the robot’s ability to identify and locate a single target. Targets were randomly distributed within a 5-meter radius to test adaptability to different spatial layouts.

\begin{table}[htb]
\centering
\caption{Experimental results under single-command tasks.}
\label{tab:Simple Instructions}
\renewcommand{\arraystretch}{1.3} 
\setlength{\tabcolsep}{10pt} 
\resizebox{1\columnwidth}{!}{
\begin{tabular}{l|cccc|cccc}
\hline
\textbf{Method} & \multicolumn{4}{c|}{\textbf{Path Nav}} & \multicolumn{4}{c}{\textbf{Targeted Search Nav}} \\
\hline
                & NE $\downarrow$ & TL $\downarrow$ & SR $\uparrow$ & SPL $\uparrow$ 
                & NE $\downarrow$ & TL $\downarrow$ & SR $\uparrow$ & SPL $\uparrow$ \\
\hline
\textbf{BrainNav}  
& \textbf{0.69}  & \textbf{6.52}  & \textbf{71} & \textbf{55}  
& \textbf{0.60}  & \textbf{7.96}  & \textbf{88} & \textbf{82}  \\
MapGPT~\cite{chen-etal-2024-mapgpt} 
& 2.95 & 4.52 & 14 & 6 
& 3.27 & 3.25 & 0 & 0 \\
\hline
\end{tabular}
}
\end{table}

As shown in Table~\ref{tab:Simple Instructions}, BrainNav achieves a success rate of 88\% in target-search navigation tasks, significantly outperforming the baseline method MapGPT, which achieves 0\%. These tasks rely on the robot's ability to recognize and localize target objects. BrainNav extracts key scene features using the Visual Cortex Perception Engine and generates action strategies through the Prefrontal Decision Center , enabling effective target recognition and dynamic path adjustment to gradually approach the target. In contrast, MapGPT often causes the robot to terminate navigation prematurely before reaching the target.

In path navigation tasks, BrainNav achieves a 71\% success rate and an SPL of 55\%. These tasks require stronger global path planning and environmental adaptability. Although BrainNav adopts dual-map and dual-orientation strategies, the limited spatial reasoning ability of large models in static scenes may still lead to local exploration, reducing efficiency and negatively affecting SPL performance.

By comparison, MapGPT relies on idealized assumptions within simulators and lacks adaptability to real-world environments, resulting in low success rates even on simple tasks. BrainNav, on the other hand, leverages bio-inspired spatial cognition and hierarchical decision-making mechanisms to effectively reduce spatial hallucinations and significantly improve adaptability in real scenarios. Its strong performance in both target-search and path navigation tasks demonstrates its practicality and generalization capability in real-world applications.

\begin{table}[t]
\centering
\caption{Experimental results under complex instruction tasks}
\label{tab:complex instruction}
\renewcommand{\arraystretch}{1.2} 
\resizebox{\textwidth}{!}{%
\begin{tabular}{l|cccc|cccc|cccc|cccc}
\hline
\textbf{Method} & \multicolumn{4}{c|}{\textbf{Multi-step}} & \multicolumn{4}{c|}{\textbf{Avoidance}} & \multicolumn{4}{c|}{\textbf{Multi-target}} & \multicolumn{4}{c}{\textbf{Interaction}} \\
\hline
                & NE $\downarrow$ & TL $\downarrow$ & SR $\uparrow$ & SPL $\uparrow$ 
                & NE $\downarrow$ & TL $\downarrow$ & SR $\uparrow$ & SPL $\uparrow$ 
                & NE $\downarrow$ & TL $\downarrow$ & SR $\uparrow$ & SPL $\uparrow$ 
                & NE $\downarrow$ & TL $\downarrow$ & SR $\uparrow$ & SPL $\uparrow$ \\
\hline
\textbf{BrainNav}   & \textbf{0.53}    & \textbf{19.67}   & \textbf{67}    & \textbf{55}    
                    & \textbf{0.32}    & \textbf{11.17}   & \textbf{87.5}   & \textbf{72}     
                    & \textbf{2.95}    & \textbf{16.78}    & \textbf{35}     & \textbf{28.5}   
                    & \textbf{1.74}    & \textbf{7.45}    & \textbf{25}     & \textbf{26.4}   \\
MapGPT\cite{chen-etal-2024-mapgpt}          & 3.71            & 9.21            & 0          & 0           
                    & 4.43            & 10.60            & 0              & 0              
                    & 3.78            & 12.5            & 0           & 0           
                    & 1.13            & 13.78            & 8           & 3           \\
\hline
\end{tabular}
}
\end{table}

\subsubsection{Complex Instruction Tasks}


Complex instruction tasks require robots to complete missions through exploration and path adjustment without directly observing the target. These tasks mainly include multi-step navigation, dynamic obstacle avoidance, multi-target search, and environment interaction. As shown in Table~\ref{tab:complex instruction}, BrainNav outperforms MapGPT across all complex tasks. BrainNav utilizes bio-inspired cognitive mechanisms to construct dual maps in real time, enabling long-distance navigation. Notably, in dynamic obstacle avoidance tasks, it achieves an 87.5\% success rate relying solely on vision and hierarchical decision modules, without radar assistance. In multi-target search tasks, the system leverages memory modules and strategy planning to effectively complete sequential target recognition, although occasional path redundancy occurs when targets are dispersed. Performance in environment interaction tasks is weaker (success rate only 25\%), mainly limited by the randomness of human movements, indicating room for improvement in simultaneously handling following and obstacle avoidance.

\subsubsection{Backtracking Rate}

\begin{table}[htb]
    \centering
    \caption{Backtracking rate and backtracking success correction rate comparison}
    \label{tab:backtracking_comparison}
    \renewcommand{\arraystretch}{1.1} 
    \setlength{\tabcolsep}{10pt}      
    \begin{tabular}{l|c|c}  
    \hline
    \textbf{Method} & \textbf{BTR} & \textbf{BSCR} \\
    \hline
    \textbf{BrainNav} & \textbf{10\%} & \textbf{73.3\%} \\
    MapGPT\cite{chen-etal-2024-mapgpt} & 75\% & 0\% \\
    \hline
    \end{tabular}
\end{table}

Backtracking rate measures how frequently the robot revisits previously visited locations during navigation. Table~\ref{tab:backtracking_comparison} shows that BrainNav’s backtracking rate is only 10\%, much lower than MapGPT’s 75\%, indicating more stable path planning. Even when backtracking occurs, BrainNav maintains a 73.3\% success correction rate, thanks to its hippocampal memory center dynamically updating and adjusting environmental information. In contrast, MapGPT frequently backtracks in real environments due to spatial hallucinations and struggles to effectively correct its path.

\subsection{Ablation Study}
In the ablation experiments, we sequentially removed the five core modules of BrainNav to evaluate their impact on navigation performance. Removing the parietal spatial builder reduced the success rate (SR) to 0\%, indicating that relying on a single-map system causes the robot’s self-coordinate confusion and severely impairs path planning ability. Without the visual cortex perception engine, SR dropped to 0\%, showing that without visual information, the robot loses environmental perception and completely fails to navigate. Removing the prefrontal decision center decreased SR to 44.6\%, due to fragmented navigation strategies and reduced path efficiency. After removing the hippocampal memory center, SR was 23.7\%; lacking historical memory caused the robot to make many detours, significantly lowering navigation efficiency but still enabling it to find targets probabilistically. When the cerebellum motor execution unit was removed, SR was 0\%, as the robot could not perform any actions and remained stuck in place. These results fully validate the critical roles of each module in precise perception, efficient planning, and stable execution, demonstrating the importance of BrainNav’s multi-module collaborative design in enhancing navigation performance.

\subsection{Case Study}

\begin{figure}[htb] 
    \centering
    \includegraphics[width=1\textwidth]{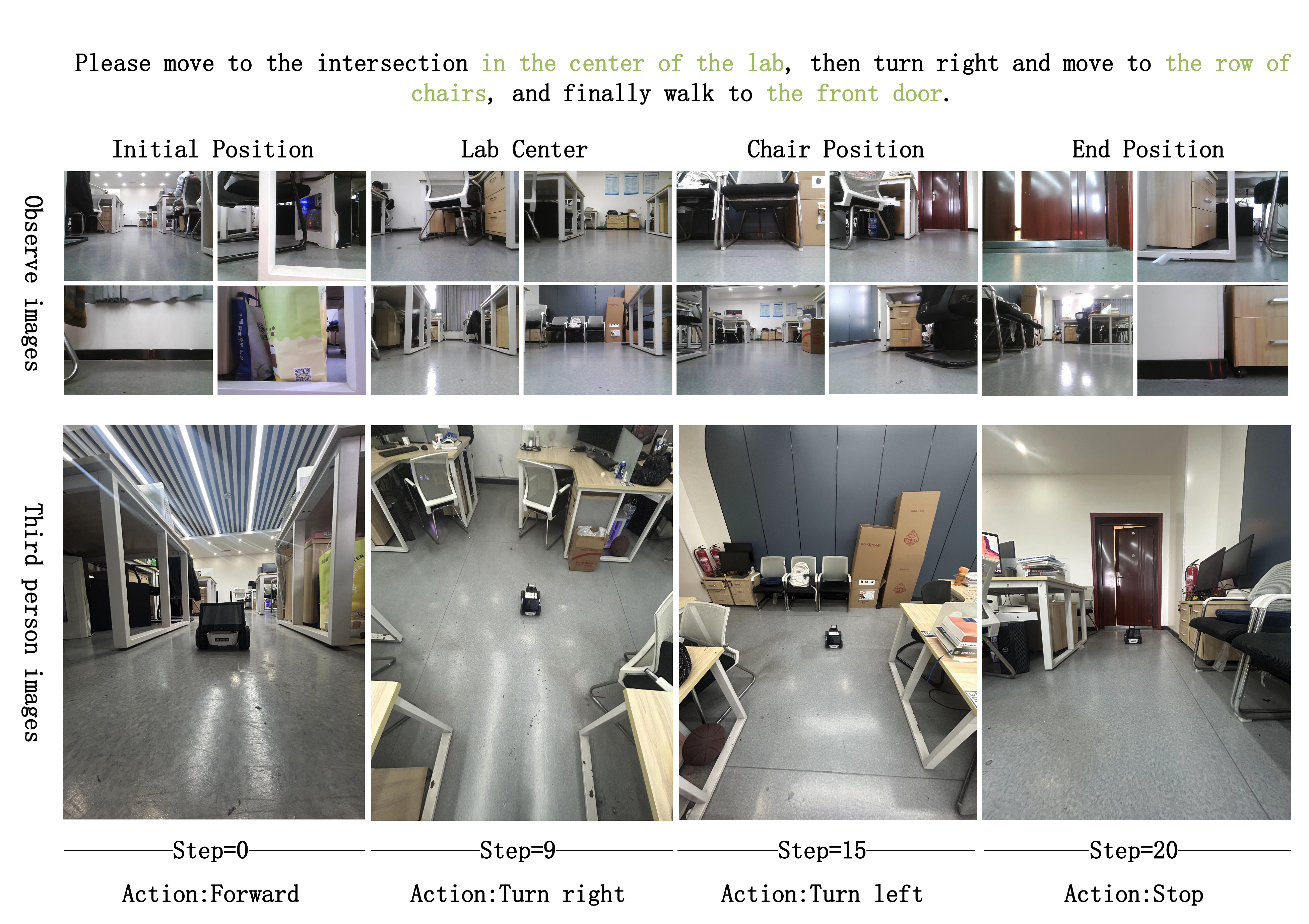} 
    \caption{
        A successful case of complex instructions in real-world exploration. We provide both the robot's perspective and a third-person perspective, demonstrating its effective multi-task navigation capabilities.
    }
    \label{fig:case} 
\end{figure}

In Figure \ref{fig:case}, we show the robot executing the instruction: “First move to the intersection at the center of the lab, then turn right to a row of chairs, and finally go to the door.” This task requires multiple turns in a narrow space and precise identification of key locations such as the intersection, cabinet, and door.

When the agent receives the navigation instruction, if both the instruction and environment match a previous task exactly, the system directly invokes the Cerebellar Motion Execution Unit to reuse the path stored in the Hippocampal Memory Hub for fast execution. Notably, BrainNav performs zero-shot navigation in unknown environments but stores experienced paths to support future reuse in the same scene and task.
Since this is the first exploration of the lab, the system activates the Visual Cortex Perception Engine, which uses the agent’s visual sensors to capture real-time images and extract object features (e.g., walls, red door, chairs, white backpack) and their relative distances to the agent. These features build semantic environmental information, providing foundational data for subsequent path planning.
The Parietal Spatial Constructor then builds and updates a dual-map (coordinate and topological) of the lab, offering multi-level environment representation to support efficient path planning. The Prefrontal Decision Center generates stepwise navigation strategies based on the instruction—moving from the start point to the lab center intersection, turning right to the cabinet, then left to the door—ensuring gradual task completion. It also converts high-level navigation instructions into specific action commands.
Finally, the Cerebellar Motion Execution Unit performs precise motion control, enabling the agent to execute multiple maneuvers in narrow spaces and accurately reach the target locations. After several iterations, the agent successfully completes the navigation task, validating BrainNav’s navigation capability and task efficiency in complex environments.

\section{Conclusion}

This paper proposes the bio-inspired hierarchical decision framework BrainNav, which integrates neuroscience-inspired multi-module collaboration to achieve state-of-the-art performance in VLN tasks. Core innovations include the fusion of absolute and relative directions to alleviate spatial hallucinations, and the use of dual-map to enhance spatial cognition robustness. Current limitations include decreased success rates in multi-object referring tasks and instability in navigation within dynamic obstacle environments. Future work will focus on optimizing multimodal semantic understanding and adaptability to dynamic environments.







\bibliography{references}

\appendix
\section*{Appendix}
\addcontentsline{toc}{section}{Appendix}

\subsection{Prompts}

\begin{figure*}[htb]
    \centering
    \includegraphics[width=\textwidth]{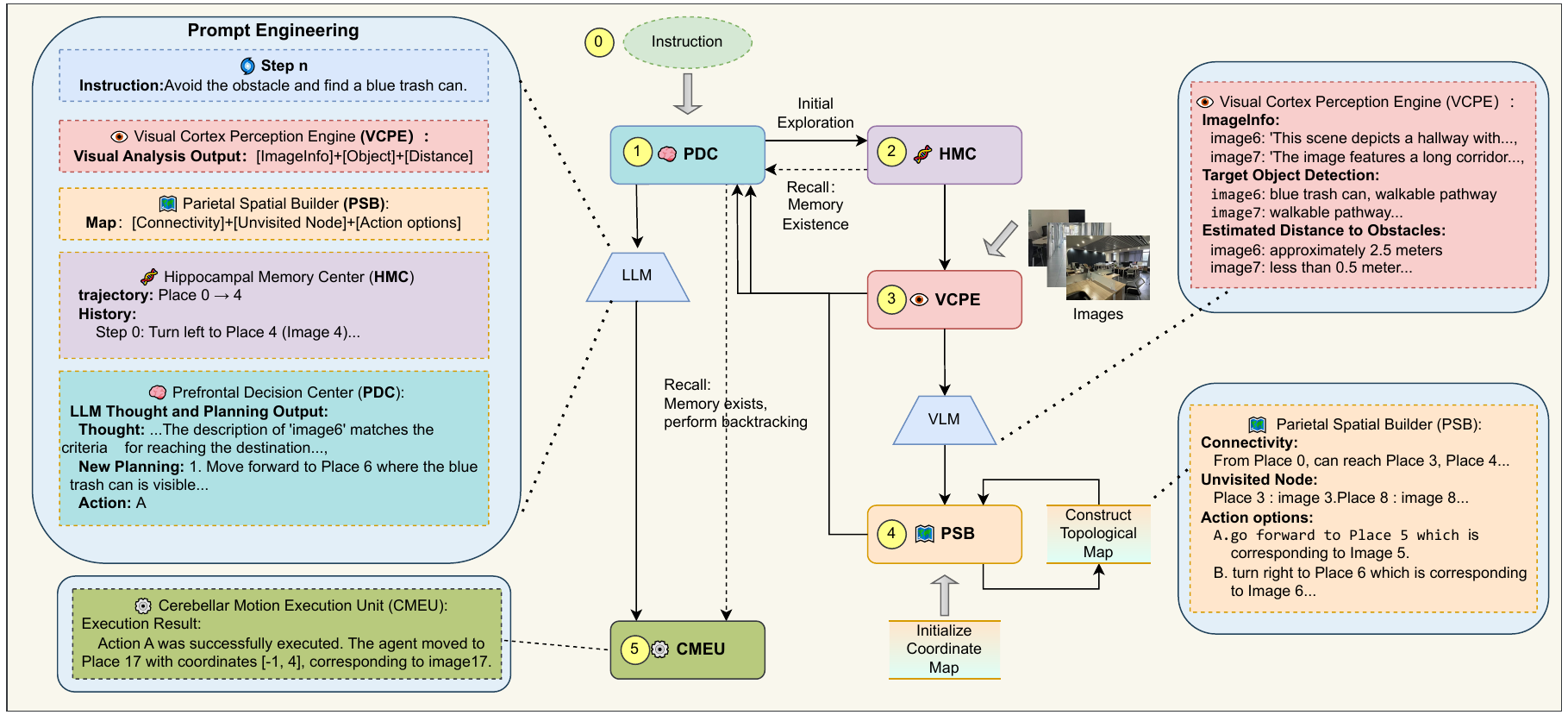}
    \caption{
        Prompt Architecture. The central portion shows the BrainNav architecture, with corresponding prompts displayed on the left and right sides.
    }
    \label{fig:prompt}
\end{figure*}

Figure~\ref{fig:prompt} presents an example illustrating the dialogue and navigation module outputs.  
In Step 2, the robot receives the instruction: “Avoid the obstacle and find a blue trash can.” The Visual Cortex Perception Engine (VCPE) first processes the visual input, extracting image features, identifying objects, and estimating distances to establish the perceptual foundation for navigation. The Parietal Spatial Builder (PSB) maintains a topological map of the environment, indicating that the current location, Place~4, connects to Place~6 and Place~8, and it lists unvisited nodes along with multiple possible actions.

The Hippocampal Memory Center (HMC) records the robot’s previous trajectory and navigation history, helping to avoid redundant paths and spatial hallucinations. Based on inputs from perception and memory, the Prefrontal Decision Center (PDC) uses the language model to reason that “image6” matches the target description, plans to move to Place~6, and verifies the distance to the target. It then decides to execute Action~A (move forward to Place~6). Finally, the Cerebellar Motion Execution Unit (CMEU) carries out the action and relocates the robot, completing the navigation step.  

This process forms a closed-loop system of perception, memory, decision-making, and execution, reflecting biological brain mechanisms of spatial navigation.

\begin{figure*}[htb]
    \centering
    \includegraphics[width=\textwidth]{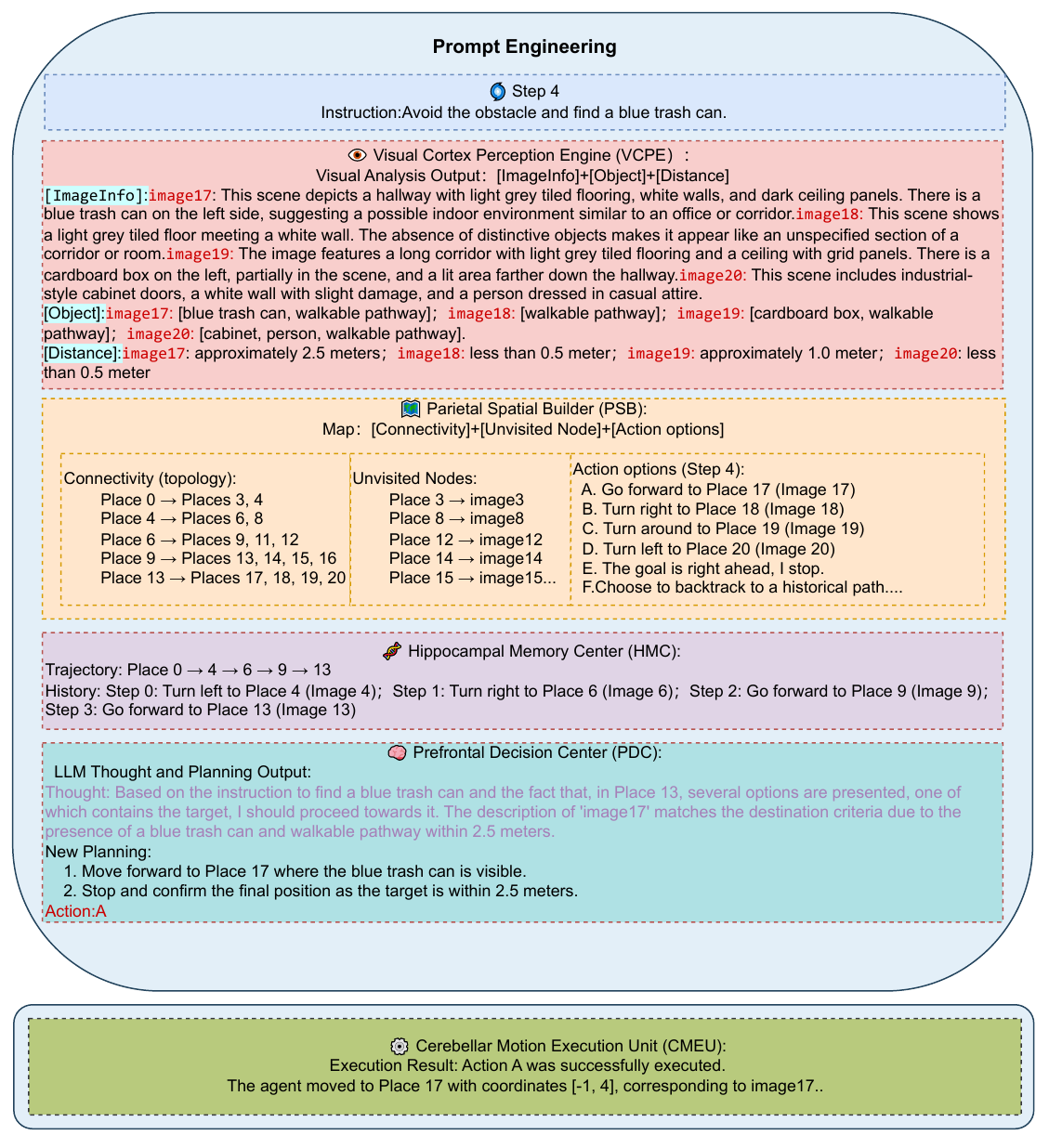}
    \caption{Prompt Architecture – Step 4.}
    \label{fig:prompt4}
\end{figure*}

\subsection{Command Categories}

\begin{table*}[t]
    \centering
    \caption{Commands categorized into simple (1–4) and complex (5–10), corresponding to different navigation tasks.}
    \label{tab:Command}
    \renewcommand{\arraystretch}{1.15}
    \setlength{\tabcolsep}{6pt}
    \begin{tabular}{c|c|p{10cm}}
        \hline
        \textbf{No.} & \textbf{Category} & \textbf{Instruction} \\
        \hline
        1 & Targeted Search Navigation & Find a blue trash can. \\
        \hline
        2 & Targeted Search Navigation & Walk toward the table where the flowerpot is placed. \\
        \hline
        3 & Path Navigation & Walk to the cabinet on the left front side. \\
        \hline
        4 & Path Navigation & Move in between the box and the chair. \\
        \hline
        5 & Multi-target Navigation & Walk down the hallway to the blue trash bin, then look for a pot of green plants, and finally stop by the staircase. \\
        \hline
        6 & Multi-target Navigation & First walk to the side of the window, then proceed to a chair. \\
        \hline
        7 & Multi-step Navigation & Proceed along the corridor, pass by a room, then turn right and enter the laboratory's main door. \\
        \hline
        8 & Multi-step Navigation & Follow these steps: move forward, move backward, turn left, turn right, and stop. \\
        \hline
        9 & Barrier Avoidance Navigation & Turn left to avoid the obstacle in order to find the blue trash can. \\
        \hline
        10 & Interactive Navigation & Find a person and follow them. \\
        \hline
    \end{tabular}
\end{table*}

Some commands are shown as icons, as illustrated in Table~\ref{tab:Command}.

\end{document}